# Time Series Anomaly Detection

## Detection of Anomalous Drops with Limited Features and Sparse Examples in Noisy Highly Periodic Data

Dominique T. Shipmon, Jason M. Gurevitch, Paolo M. Piselli, Steve Edwards

*Google, Inc.*
*Cambridge, MA, USA*
`{dshipmon205, jasongu927}@gmail.com`[1]
`{ppiselli,stedwards}@google.com`

*Abstract*—**Google uses continuous streams of data from industry partners in order to deliver accurate results to users. Unexpected drops in traffic can be an indication of an underlying issue and may be an early warning that remedial action may be necessary. Detecting such drops is non-trivial because streams are variable and noisy, with roughly regular spikes (in many different shapes) in traffic data. We investigated the question of whether or not we can predict anomalies in these data streams. Our goal is to utilize Machine Learning and statistical approaches to classify anomalous drops in periodic, but noisy, traffic patterns. Since we do not have a large body of labeled examples to directly apply supervised learning for anomaly classification, we approached the problem in two parts. First we used TensorFlow to train our various models including DNNs, RNNs, and LSTMs to perform regression and predict the expected value in the time series. Secondly we created anomaly detection rules that compared the actual values to predicted values. Since the problem requires finding sustained anomalies, rather than just short delays or momentary inactivity in the data, our two detection methods focused on continuous sections of activity rather than just single points. We tried multiple combinations of our models and rules and found that using the intersection of our two anomaly detection methods proved to be an effective method of detecting anomalies on almost all of our models. In the process we also found that not all data fell within our experimental assumptions, as one data stream had no periodicity, and therefore no time based model could predict it.**

## I. Introduction

### A. About the Data

We looked at 14 different sets of data that were saved at 5 minute intervals. This means that that each hour had 12 data points, and each day had 288. Our goal is to detect anomalies such as absences of daily traffic bursts on unexpected decreases in the throughput volume, while not detecting anomalies on regular low -- or even 0-- values. The only two attributes of this dataset were a unix timestamp and the bytes per second value, both sampled every 5 minutes. Although there are no accurate markings of anomalies within this data, there does not mean that they do not exist, only that we had no information of the at the beginning of this project.

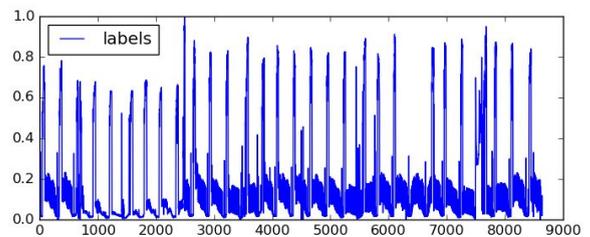

### B. About the Problem

When there are clear labels for anomalous data a binary classifier can be built predict anomalies and non-anomalous points. For these problems there exist a plethora of techniques to choose from, including clustering analysis, isolations forests, and classifiers built using artificial neural networks. These have all shown promise in the field of anomaly detection [1]. The first two of these techniques not only require labels for training, but also are most effective when there are many features. For these reasons they are not as useful on time series data, especially when there are not other features that we have access to.

Neural networks can effectively predict periodic time series data, as can simpler techniques such as Fourier series. However, because what counts as an anomaly can vary based on the data, each problem potentially requires its own model. Some problems have many features to work with or need to never give a false positive, while others, like ours, have few features and look for continuous, rather than point anomalies.

We trained a separate model for each data stream since we found no significant correlation between each of the data streams. Our models returned predictions that we compared to the actual data stream values using our anomaly detection rules to determine if the current point is anomalous. Although all of our models and metrics are done in an offline environment, they are all approaches that are adaptable to online training, predicting, and anomaly detection. The adaptability of our models is a result of our machine learning models and anomaly detection rules only requiring either the past or current points. Additionally, to make sure our models were effective we compared them to a threshold model, which would constantly predict a very low value.

Since our models are only predicting a point through regression, we need a defined rule to determine whether a point is anomalous or not by comparing the prediction and actual value. The intuitive approach for comparing the predictions actual value to determine whether a point is anomalous or not is by simply calculating the euclidean distance and setting a threshold. However, this could lead to numerous false positives especially with noisy data. Therefore, we implemented two different detection rules for anomaly detection: one using an accumulator to detect continuous outages and another using a probabilistic approach as outlined by Ahmad and Purdy [2]. Both methods attempt to mitigate high false positive rates by avoiding a simple threshold technique applied to one datapoint. The accumulator method works by having a metric defined to determine a local outage, and then increment the counter for every outage and decrement it for every non-anomalous value. The statistical method based on the Numenta's formulation [2] worked by comparing short term variance to long term variance, meaning that it would also adjust to the data in addition to our model [2]. We found both of these methods to work similarly well; they both fired at almost exactly the same points. Despite this, each offers theoretical and practical approaches, particularly in potential future experiments, since the accumulator method is easily modifiable and controllable, and the statistical method gives a percentage chance of being an outlier, which could be multiplied between models to give a more robust score [2].

We found that for this problem it was the anomaly detection method that mattered more than the model itself, since our anomaly rules were able to be quite effective on most of our models.

## II. Related Work

Anomaly detection is actively and heavily researched [3][4]. While classification techniques are a popular approach to solve anomaly detection, it is unrealistic to expect to always have a dataset with a sufficient and diverse set of labeled anomalies [1]. Statistical and regression techniques seem more promising in these cases. Netflix recently released their solution for anomaly detection in big data using Robust Principle Component Analysis [5]. Netflix's solution proved to be successful however, their statistical approach relies on high cardinality data sets to compute a low rank approximation which limits its applications [5]. Twitter also released an approach to anomaly detection which considers seasonality using the Seasonal Hybrid Extreme Studentized Deviate test (S-H-ESD) [6]. The technique employed by Twitter is promising as it is suited for breakout anomalies, not just point anomalies [6]. There are many other machine learning techniques that could be used to detect anomalies, such as regression using LSTMs, RNNs, DNNs or other model types [1]. In this paper we focus on the performance of regression-based machine learning techniques on periodic time series data with limited features and few labeled examples of anomalies.

## III. Data Preprocessing and Initial Analytics

The only feature available for us to use is the unix time stamp and the only label is the number of bytes (amount of data) received. Therefore, we wanted to figure out what methods, if any, we could use to create other features that are not linearly correlated with the original timestamp feature. While a Fourier Series can only work on a single value, other machine learning approaches work better with more features. Furthermore, since we are training on labels, it cannot be an input feature, which means without any feature engineering there is only one feature for our model to use.

We decided to see if any form of composite data streams would make sense as a feature, such as the average features value of all streams or using the principal components to decrease it to a few significant streams. However, we found that almost none of the streams had any linear correlation to each other through calculating the covariance matrix, meaning that a composite based on principal component analysis would not add information

As a result, all of our features had to come from just the single data stream we were working with. For each of our network models we converted the unix timestamp into a weekday, hour, and minutes feature. Then we decomposed the time features into multiple, non-linearly correlated features by converting hours and weekday into one hot encodings, while adding a linear minutes feature in the hopes that it could help the model learn better by creating more features with more complicated relations. We also decided to experiment with using the derivative of past labels as a feature. The only variance possible without using the derivative is within the largest timeframe fed into the models, which in our case was a week. This means that the same values would be predicted every week by our models. Furthermore the choice of using the derivative of past data points was driven by its ability to be used in online monitoring. Although using features computed from the values we are trying to predict is often poor practice due to overfitting, we hoped that the derivative would be weakly related enough --especially as a non-linear transform of the labels-- that the models learned time as a more important feature. When we simulated the anomaly to be a continuous value, such as 0, we found that it continue to predict spikes, although they were slightly smaller, showing that our models were not overfitting on the derivative.

We normalized the bytes per second metric to be from 0 to 1 because while the Fourier Model could work with values as high and variable as our inputs, approaches with Neural

Networks are much less effective when the values are that high. Furthermore, setting these bounds allowed us to use activation functions that transformed data to be from -1 to 1 or 0 to 1 like the hyperbolic tangent function (tanh) and the sigmoid function, to try more methods to optimize our models. Additionally normalizing the data made training faster even for the Fourier Model, so we used normalized data for all of our models.

## IV. DETECTION RULES

We experimented with two types of detection rules, an accumulator method and a gaussian tail probability method. Through our experimentation we found that with each data stream either anomaly detection rule was more effective. Therefore we found that a hybrid model which is simply the intersection of the two models is even more effective at identifying anomalous points. We found instances where each detection rule worked better than the other, therefore we used the intersection of the accumulator and the gaussian tail probability rules.

### A. Accumulator

The goal of the accumulator rule was to require multiple point anomalies to occur in a short period of time before signalling a sustained anomaly. We tried two different rules for how a point anomaly is detected, and then tested both of them as part of the accumulator rule. This rule involved a counter that would grow as point anomalies are detected, and shrink in cases where the predicted value is correct. The goal of this is to prevent noise that a local anomaly detection algorithm would incur, by requiring multiple anomalies in a short timeframe to cause it to reach a threshold for signaling. For every local anomaly the accumulator grows by one, and for every non-anomalous value it shrinks by two so that pure noise causes the accumulator to shrink. The accumulator is capped between 0 and 1.5x the value of the threshold for signalling, although these numbers were chosen based on testing and other optimal values could be possible.

#### 1) Threshold

This algorithm defined a local anomaly as any value where the actual value is more than a given delta below the expected value, when the actual value is greater than a hardcoded threshold which says any value above it is not anomalous. This was just a simple algorithm, again parametric in nature where the values we chose worked for our dataset, but again may not be optimal for every possible one.

#### 2) Variance Based

We tried using local variance to determine outages by defining an outage as any value that is outside of 20 times the rolling variance from the current prediction, so that noisy areas would allow for more noise in anomalies. However, this method proved to be slightly less effective than the simple threshold.

Because we found the threshold method to be more effective on our data, we used that in all of our anomaly detection for this paper. We observed that the accumulator rules frequently identified false positives after peaks due to an offset in the models inference compared to the ground-truth. This offset is likely due to training on a previous months where changes in periodicity could occur gradually into the next evaluation month. In an attempt to abate these false positives, we added a parameter for 'peak values' that would decrement the accumulator by three following a peak, and allowing the accumulator to go below 0 (down to a negative the threshold). However, we only wanted this to affect prediction immediately after peaks, so the accumulator would decay back to 0 if more non-anomalous data points are found, effectively preventing it from predicting an anomaly immediately after a peak.

For all of our models we defined the threshold for a non-anomalous value at 0.3, a peak value at 0.35, had a an accumulator threshold at 15 and had the delta for a local outage at 0.1

### B. Gaussian Tail Probability

The second anomaly detection rule that we decided to test for our data is the Gaussian tail probability rule defined by Numenta. Numenta's rule first requires the computation of a raw anomaly score between the inference value and the ground truth [2]. The raw anomaly score calculation is simply the difference between the inference and ground-truth, however, points above inference are not considered as we are only concerned with a lack of activity. Therefore, our raw anomaly score is as follows, where $f(x_t)$ is the prediction at time $t$ and $a_t$ is the ground-truth:

$$s_t = max(f(x_t) - a_t, \ 0)$$

The series of resulting raw anomaly scores are used to calculate the rolling mean and variance. The rolling mean has two windows where the length of W2 < W1. The two rolling means and variance are used to calculate the tail probability which is the final anomaly likelihood score.

## V. PREDICTION MODELS

We started with testing each of our models on two artificial sets that we created, before using them on real data. One of the artificial datasets was the sine function with a period of 288 data points to see if the models can predict periodic data of the same period as our data, and the other was a stepwise function based on the sine function with the same period, to see if the models could predict periodic data that changed rules at certain regular intervals.

Since our data was a continuous stream we broke each stream into months, and trained on April data and tested on May data. 13 of the 14 datasets we looked at were periodic in nature, although there was variance in exactly when spikes would start, in addition to the exact values of these spikes and the amount of noise present.

For each of our models we evaluated them using both of our anomaly detection rules, the accumulator method, Numenta's tail probability method [2], as well as investigating the effectiveness of the intersection of both methods. Additionally, each model, except for the fourier model was trained using the Adam Optimizer, where the fourier model is trained using the Adagrad optimizer.

### A. Hyperparameters

All hyperparameters were experimentally chosen based on effective models. We attempted to use hyperparameter optimization based on mean squared error loss, however, this did not provide useful models, consequently, we continued using our experimentally chosen hyperparameters.

### B. Baseline (Threshold model)

Our most simple model was based on what is already in place in the system, which is just a simple threshold. We used this both as a base metric for our experiments, but also to see how effective the anomaly detection algorithms could be if only applied to what was currently in place. The value for this threshold was hardcoded to 0.065, since the goal was not to minimize distance from all data points, but rather to have a value the data should not go below.

Although the accumulator model could be tweaked to be somewhat effective on this model, at least for detecting when data is abnormally low, the sliding windows method was not as effective at detecting anomalies for this model. This could in part be because variance of the error is only based on the labels, and not the labels and predictions in this model, so it cannot be as robust as it is supposed to be. The results for the baseline are in the following tables. Since our threshold was so low we needed to change the value of the delta for our accumulator model to 0.05 from 0.1 to detect a local outage.

| Baseline Training and Validation Loss | Training | Validation |
|---|---|---|
| MSE Loss | N/A | 0.056214 |

| Confusion Matrices for Anomaly Rules Using Baseline | | | |
|---|---|---|---|
| Anomaly Rule | Accumulator | Tail Probability | Intersection |
| True Negatives | 7833 | 8211 | 8211 |

| False Negatives | 183 | 366 | 366 |
| True Positives | 246 | 63 | 63 |
| False Positives | 378 | 0 | 0 |

### C. Fourier Series

As another fairly simplistic baseline model, we created a Fourier model under the assumption that most of our data has some level of periodicity to it. Rather than using a sum sines and cosines our formulation rather makes the phase parametric and summed with the harmonic as a argument of the sine. The formulation for the Fourier series used is as follows, where, $a_n$ is the amplitude, $\varphi$ is the phase and $\phi$ is the number of harmonics:

$$f(x) = \sum_{n=0}^{\phi} a_n \, sin(n + \varphi_n)$$

The loss function is defined as Mean Squared Error (MSE) and the function is optimized using Adagrad Optimizer. For the Fourier model we experimented with having the period as either a week or a day. We found that both periods were similarly effective, therefore we decided to use the one with the longer period to capture more complexities in the data. We used 448 harmonics, or 64 for each of the 7 days, and trained with a learning rate of 0.5 for 3000 steps. For it's relative simplicity the fourier attained a fairly low loss value on our data set:

| Fourier Model Training and Validation Loss | Training | Validation |
|---|---|---|
| MSE Loss | 0.0180229 | 0.045148 |

The following table lists the confusion matrices for each anomaly detection rule using the Fourier model:

| Confusion Matrices for Anomaly Rules Using Fourier Model | | | |
|---|---|---|---|
| Anomaly Rule | Accumulator | Tail Probability | Intersection |
| True Negatives | 7797 | 8009 | 8072 |
| False Negatives | 375 | 429 | 429 |
| True Positives | 54 | 0 | 0 |
| False Positives | 414 | 202 | 139 |

### D. Deep Neural Network (DNN)

For our Deep Neural Network we used the 'DNNRegressor' within the TFLearn API. The TFLearn DNN Regressor is a fully connected feed-forward model that is connected with Relu6 activation function (Relu6 is just a standard relu that caps at 6) [7]. Our network has 10 layers of 200 neurons each, and was trained for 1200 steps with a batch size of 200 and a learning rate of 0.0001 with the Adam optimizer.

This model proved quite effective and stable, which could in part be due to other optimizations performed by the TFLearn implementation of the model.

| DNN Training and Validation Loss | Training | Validation |
|---|---|---|
| MSE Loss | 0.0688874 | 0.033702 |

The following table are the confusion matrices for each anomaly detection rule using the DNN:

| Confusion Matrices for Anomaly Rules Using DNN | | | |
|---|---|---|---|
| Anomaly Rule | Accumulator | Tail Probability | Intersection |
| True Negatives | 7751 | 8003 | 8077 |
| False Negatives | 348 | 410 | 415 |
| True Positives | 81 | 19 | 14 |
| False Positives | 460 | 208 | 134 |

### E. Recurrent Neural Network (RNN)

Unlike a feed-forward DNN, an RNN contains recurrent loops where the cells output state is feed back into the input state. Such recurrent connections give RNNs the ability to have information persistence or a temporal state, therefore forming what is short term memory [8]. We decided to experiment with RNNs as our intuition is that the short term memory nature of the RNN lends itself perfectly to using past information in the time series to make improved inference. We choose to use a 75 unit hidden size, 10 layer deep RNN, with a single linear output layer. The RNN used Exponential Linear Units (ELU) activations and trained with a learning rate of 0.0001 and batch size of 200 at 2500 steps. Despite the RNN being able to make inference temporally due to its recurrent nature, we did not find a significant increase in performance to

that of the DNN. The following is a table of the RNN's loss for both testing and evaluation:

| RNN Training and Validation Loss | Training | Validation |
|---|---|---|
| MSE Loss | 0.00245542 | 0.04142 |

| Confusion Matrices for Anomaly Rules Using RNN | | | |
|---|---|---|---|
| Anomaly Rule | Accumulator | Tail Probability | Intersection |
| True Negatives | 6889 | 8056 | 8100 |
| False Negatives | 409 | 416 | 418 |
| True Positives | 20 | 13 | 11 |
| False Positives | 1322 | 155 | 111 |

### F. Long Short-Term Memory (LSTM)

An LSTM is simply another form of RNN where rather than a simple recurrent loop at reach recurrent cell, the LSTM introduces a more complex cell architecture for more accurately maintaining memory of important correlations [9]. We use a 70 hidden layer size and 10 layer deep LSTM, and as with the RNN a single linear output layer. Our LSTM used ELU activations and is trained with a learning rate of 0.001 and a batch size of 200 at 2500 steps. Furthermore, we use the standard non-peephole LSTM cell implementation. Similar to the RNN, the LSTM, despite the added complexity of temporal memory, did not perform significantly better than the DNN. The following is a table of the LSTM's loss for both testing and evaluation:

| LSTM Training and Validation Loss | Training | Validation |
|---|---|---|
| MSE Loss | 0.000980715 | 0.037625 |

| Confusion Matrices for Anomaly Rules Using LSTM | | | |
|---|---|---|---|
| Anomaly Rule | Accumulator | Tail Probability | Intersection |
| True Negatives | 7801 | 7951 | 8083 |
| False Negatives | 388 | 419 | 420 |
| True Positives | 41 | 10 | 9 |
| False Positives | 410 | 260 | 128 |

## VI. Experiment

### A. Anomaly Simulation

We first tested our models by simulating a fake anomaly to make sure they could detect something. However, since current anomaly detection methods for the data are poor we do not know what anomalies should look like, so we created a function that would create random noise around a specific value during the duration of the simulated anomaly. However, as we were testing on our data it became evident that there was an observed anomaly in late May in one of our streams. Because of this we utilized the May stream for our metrics, since we believed that anomaly to be more important to detect than our simulated ones, although we wanted to make sure our models could detect both different types. Since this anomaly was in the form of it purely missing a peak, while still sending normal data, it is hard to numerically discern exactly where the anomaly starts and ends for evaluation purposes. Therefore, we estimated those values based on our own analysis. Furthermore, these are just two examples of anomalies, but the fact that our models could pick up both types showed that they can be robust in detecting multiple types of anomalous data.

### B. Test/Train Split

For our testing and training validation process we choose to use a new time frame (a different month) rather than randomly selecting points of test data. The rationale behind this was that it would provide a realistic environment for testing the anomaly detection rules with the test predictions as they rely on previous contiguous data points. Furthermore, testing in new temporal sequences is desired for models like the LSTM and RNN which have temporal awareness.

## VII. Findings

The numerical results form the confusion matrix show two key findings. The first is that in every instance, the intersection between the accumulator method and the tail probability method reduced the amount of false positives flagged by the model. However, since the intersection makes a tighter bound around anomalous regions, the true positive rate is also decreased. The second important finding is that there is a very small difference between all of our neural network models with regards to their detection confusion matrices. This suggests that the most important factors are that the model can fit to nonlinear data and has online learning capabilities (to adjust to long term changes in the streaming time series). Furthermore, an important factor is the anomaly detection rule itself, as each neural network had a low loss and fit the test and validation data very well, the biggest difference being between the anomaly detection rules and not the models. Finally, it should be noted that while some models did not numerically perform as desired within the bounds of the labeled anomalies, some models like the Fourier model, marked anomalies very close to the labeled anomalies. These results indicatet that observed model performance is also reliant on ground-truth labeling guidelines.

The numerical results for each model also show important results both in anomaly detection performance and training validity. For training validity, the evaluation and training MSE scores demonstrate that overfitting is not occurring and the model is making a reasonable generalisation about the data as the validation scores are not significantly higher.

While the numeric data didn't quite show how effective our models were compared to each other, the graphical data (shown in the Figures section) proved to be a very effective way to gauge which models were effective and the differences between them. This was in part because the outage we simulated was during a period where there wasn't supposed to be significant traffic, so our models barely detected that fake outage, but all of them but the baseline were able to detect the real anomalies within this timeframe. Outside of detecting those two anomalies, we found the graphical approach to be effective because it shows sustained false positives vs frequent, but smaller false positives. From analyzing the graphs we can see that the Fourier Series model fired the fewest false positives on our evaluation data set, firing only four times outside of our two anomalous points. Aside from this however the graphical data shows that all of the models had similar effectiveness, and the small differences can both be down to the models themselves and the specific parameters used for the anomaly detection methods. Despite this the Fourier, and RNN models were all slightly more effective, while the DNN and LSTM had more false positives, specifically right after peaks. However, because this is such a specific and recurring type of false positive the accumulator detection rule can be easily modified to have more resistance to predictions after peaks, which would make these models just as effective as the other ones.

Although this paper only actively discusses our methods working on one of the streams and failing on one, it was able to successfully detect anomalies in 13 out of the 14 streams, and oftentimes exhibited improved performance on other streams, which are less erratic than the stream discussed in this paper. The one that our models could not work on was not due to a failure in our models, but instead because that stream was not periodic.

## VIII. Conclusion

For the streams that we analyzed all of our models were very effective with the anomaly detection rules that we used. While the Fourier was slightly more effective than the others, this can also be due to the data itself, where each datastream can have peculiarities that make one model work better than others. With access to more features, deep learning could provide even more accurate results. However, our results suggest that due to the limited set of features available it didn't provide significant advantages to simpler periodic models. Using two completely different anomaly detection rules, particularly one with statistical backing and another that can be easily tweaked by the user to manually remove recurring false positives allowed us to have a very robust method to detect anomalies. While these methods can be used for general purpose anomaly detection, the methods were somewhat specific to our problem because rather than looking at total distance from predictions we only looked at distance below prediction, although these methods should still work well on other data if we account for those differences. Room for further experimentation can be done by trying even more anomaly detection methods and seeing if another combination can work even better than the two we propose here. lso an optimization of our models using other metrics for hyperparameter optimization, that could either give us models with better results or less computationally expensive models with the same results.

n.b. Tail Probability method cannot predict anything during the first long window, which is a week, or 2016 points.
Blue line is the actual values, green line is the predictions, and shaded red areas are the detected anomalies

Baseline model

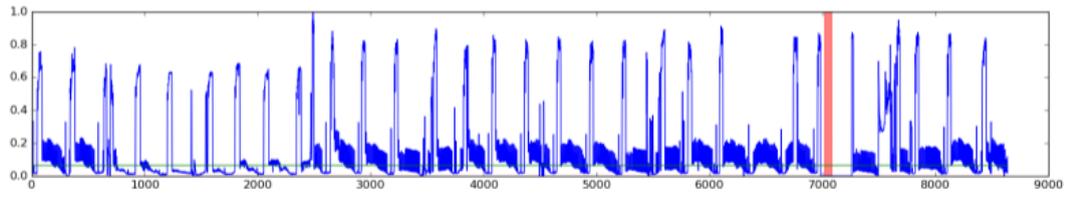

Fourier Model

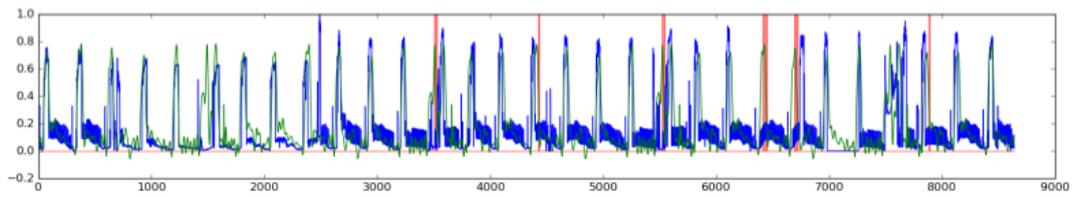

DNN model

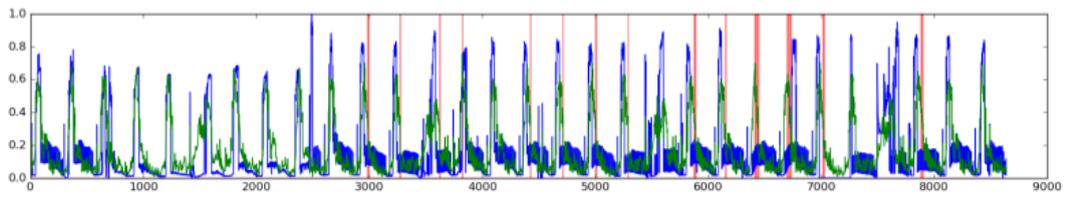

RNN model

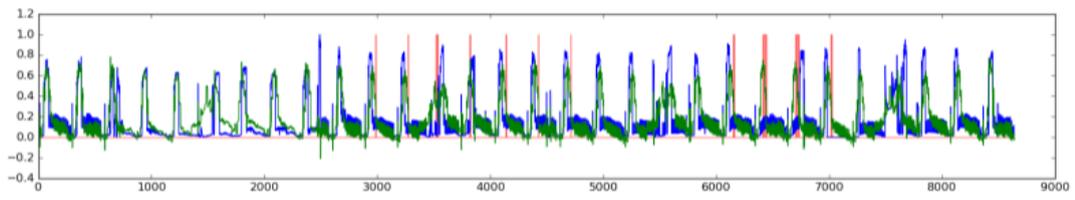

LSTM model

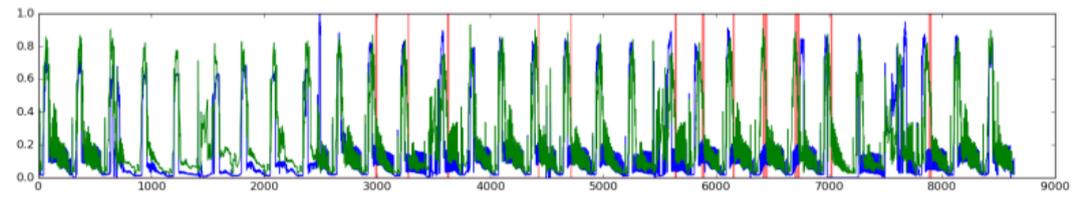

DNN on another stream

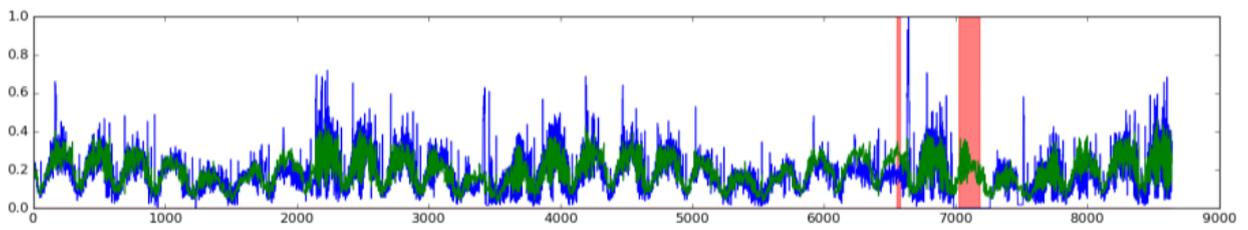